\documentclass[10pt,twocolumn,letterpaper]{article}

\usepackage[]{cvpr} 

\definecolor{cvprblue}{rgb}{0.21,0.49,0.74}
\usepackage[pagebackref,breaklinks,colorlinks,allcolors=cvprblue]{hyperref}
\usepackage{multirow}
\usepackage{placeins}
\usepackage{arydshln} 
\usepackage{hyperref}
\usepackage{newtxtext,newtxmath}

\title{GCE-Pose: Global Context Enhancement for Category-level \\ Object Pose Estimation}

\author{
Weihang Li\textsuperscript{1,2*} \ \
Hongli XU\textsuperscript{1*} \ \
Junwen Huang\textsuperscript{1,2*\dag} \ \
Hyunjun Jung\textsuperscript{1*} \\
Peter KT Yu\textsuperscript{3} \ \
Nassir Navab\textsuperscript{1,2}\ \
Benjamin Busam\textsuperscript{1,2} \\
{\textsuperscript{1}Technical University of Munich}
\,
{\textsuperscript{2}Munich Center for Machine Learning}
\,
{\textsuperscript{3}XYZ Robotics}
}

\definecolor{wen}{rgb}{0.9,0.5,0.1}

\begin{document}
\maketitle
\begin{abstract}
A key challenge in model-free category-level pose estimation is the extraction of contextual object features that generalize across varying instances within a specific category. Recent approaches leverage foundational features to capture semantic and geometry cues from data. However, these approaches fail under partial visibility. We overcome this with a first-complete-then-aggregate strategy for feature extraction utilizing class priors. In this paper, we present GCE-Pose, a method that enhances pose estimation for novel instances by integrating category-level global context prior. GCE-Pose performs semantic shape reconstruction with a proposed Semantic Shape Reconstruction (SSR) module. Given an unseen partial RGB-D object instance, our SSR module reconstructs the instance's global geometry and semantics by deforming category-specific 3D semantic prototypes through a learned deep Linear Shape Model. We further introduce a Global Context Enhanced (GCE) feature fusion module that effectively fuses features from partial RGB-D observations and the reconstructed global context. Extensive experiments validate the impact of our global context prior and the effectiveness of the GCE fusion module, demonstrating that GCE-Pose significantly outperforms existing methods on challenging real-world datasets HouseCat6D and NOCS-REAL275. Our project page is available at \url{https://colin-de.github.io/GCE-Pose/}.
\end{abstract}
\vspace{-0.5cm}    
\def\thefootnote{*}\footnotetext{Equal contribution.}\def\thefootnote{\arabic{footnote}}
\def\thefootnote{\dag}\footnotetext{Corresponding author: \texttt{junwen.huang@tum.de}
}\def\thefootnote{\arabic{footnote}}
\begin{figure}[t] 
    \hspace*{\fill} 
    \begin{minipage}[t]{0.48\textwidth} 
        \centering
        \includegraphics[width=\textwidth]{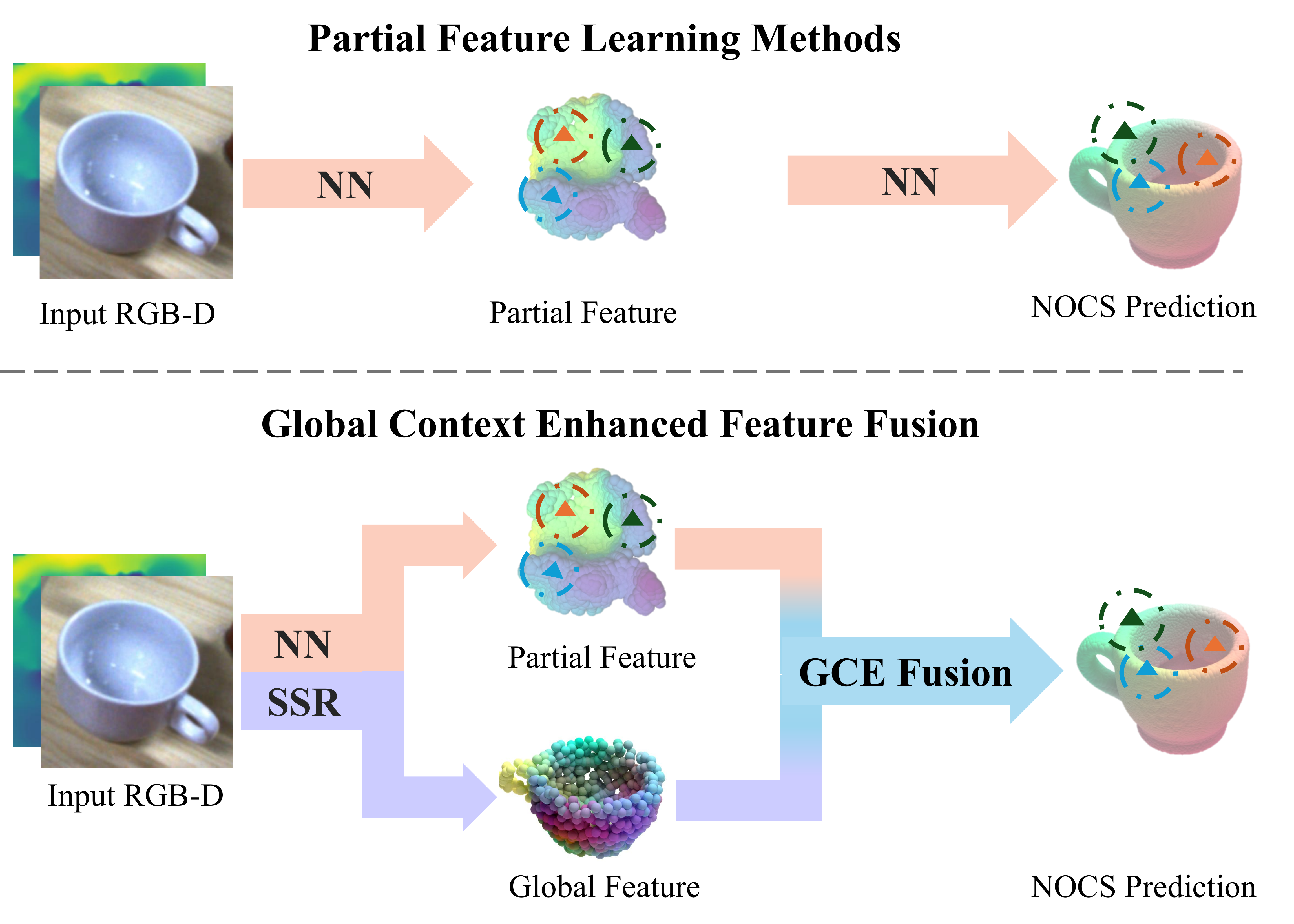} 
        \caption{Overview of the category-level Pose Estimation Pipeline: (A) Previous methods, \textit{i.g.} AG-Pose~\cite{lin2024instance} and Second Pose~\cite{chen2024secondpose}, rely on partial features extracted by a neural network (NN) to regress object poses. (B) We introduce a novel approach that leverages a semantic shape reconstruction (SSR) module for global feature extraction. This global context enhances (GCE) the mapping from partial features to NOCS features.}
        \vspace{-0.7cm}
        \label{fig:teaser}
    \end{minipage}
\end{figure}

\section{Introduction}
\label{sec:intr}
The task of object pose estimation varies according to generalization level and input modality. \textit{Instance-level pose estimation} methods~\cite{Wang_2021_GDRN,su2022zebrapose,hodan2024bop} focus on specific object instances that does not generalize to other objects, while \textit{fully unseen object pose estimation} methods~\cite{drost2010model,shugurov2022osop,huang2024matchu} are designed to handle novel objects, however requires object model as prior. Unlike aforementioned methods, \textit{category-level pose estimation} methods~\cite{nocs,lin2022category,chen2024secondpose} aim to generalize across unseen instances within a defined category that requires only an RGB(-D) image of a new instance during the inference, making the method model-free that does not require predefined object models.

Current category-level approaches primarily estimate the Normalized Object Coordinate Space (NOCS)~\cite{nocs} and employ a pose solver, such as the Umeyama algorithm~\cite{umeyama1991least}, to obtain the object pose~\cite{lin2022category}. To effectively extract category-level features from RGB (and/or depth) inputs, researchers have developed various neural network architectures to capture features from partial RGB(-D) observations. Some recent methods~\cite{ornek2025foundpose,nguyen2024gigapose,ausserlechner2024zs6d} leverage foundation models like DINOv2~\cite{oquab2023dinov2} for improved performance. Additionally, research~\cite{huang2024matchu,chen2024secondpose,caraffa2025freeze,lin2024sam,huang2024ssr} highlights the importance of combining semantic and geometric information to enhance feature robustness and distinguishability, aiding in better correspondence and pose estimation. However, category-level pose estimation being model-free and having only partially observed RGB(-D) inputs limits the extraction of global context information.
Some methods~\cite{tian2020shape,query6dof,chen2021sgpa,zhang2022rbp,lin2022category} have introduced categorical geometric shape priors to reconstruct instance models from partial input points, solving for object pose by establishing dense correspondence between partial input points and reconstructed models. However, these methods solely introduce shape priors neglecting the semantic context of the category. More recently, GS-Pose~\cite{wang2025gs} selects one instance as a reference prototype within a category and applies semantic feature matching between partial points and the reference instance. However, this design struggles with intra-class shape variations and is particularly vulnerable to noise in partial point cloud observations. 

In this work, we propose GCE-Pose, a novel approach that integrates global context incorporating both geometric and semantic cues to enhance category-level object pose estimation. We propose two major modules named Semantic Shape Reconstruction (SSR) and Global Context Enhanced (GCE) feature fusion modules to facilitate pose estimation. The SSR module is a first-complete-then-aggregate strategy that reconstructs the input partial points into a complete shape and smoothly aggregates the semantic prototype to the instance. The GCE feature fusion model is proposed to effectively fuse the reconstructed global context with local cues. The efficacy of our proposed method is confirmed by extensive evaluation on the challenging real-world datasets, achieving SOTA performance against the existing approaches. Our main contributions are as follows:

\begin{itemize}
    \item We propose GCE-Pose, a Global Context Enhancement (GCE) approach that integrates global context with both geometric and semantic cues for category-level object pose estimation.
    
    \item We introduce a Semantic Shape Reconstruction (SSR) strategy that addresses partially observed inputs by reconstructing both object geometry and semantics through learned categorical deformation prototypes.
    
    \item Extensive experiments demonstrate that our method achieves robust pose estimation even under significant shape variations and occlusions improving the generalization to unseen instances.
\end{itemize}

\begin{figure*}[t]
    \centering
    \includegraphics[width=\textwidth]{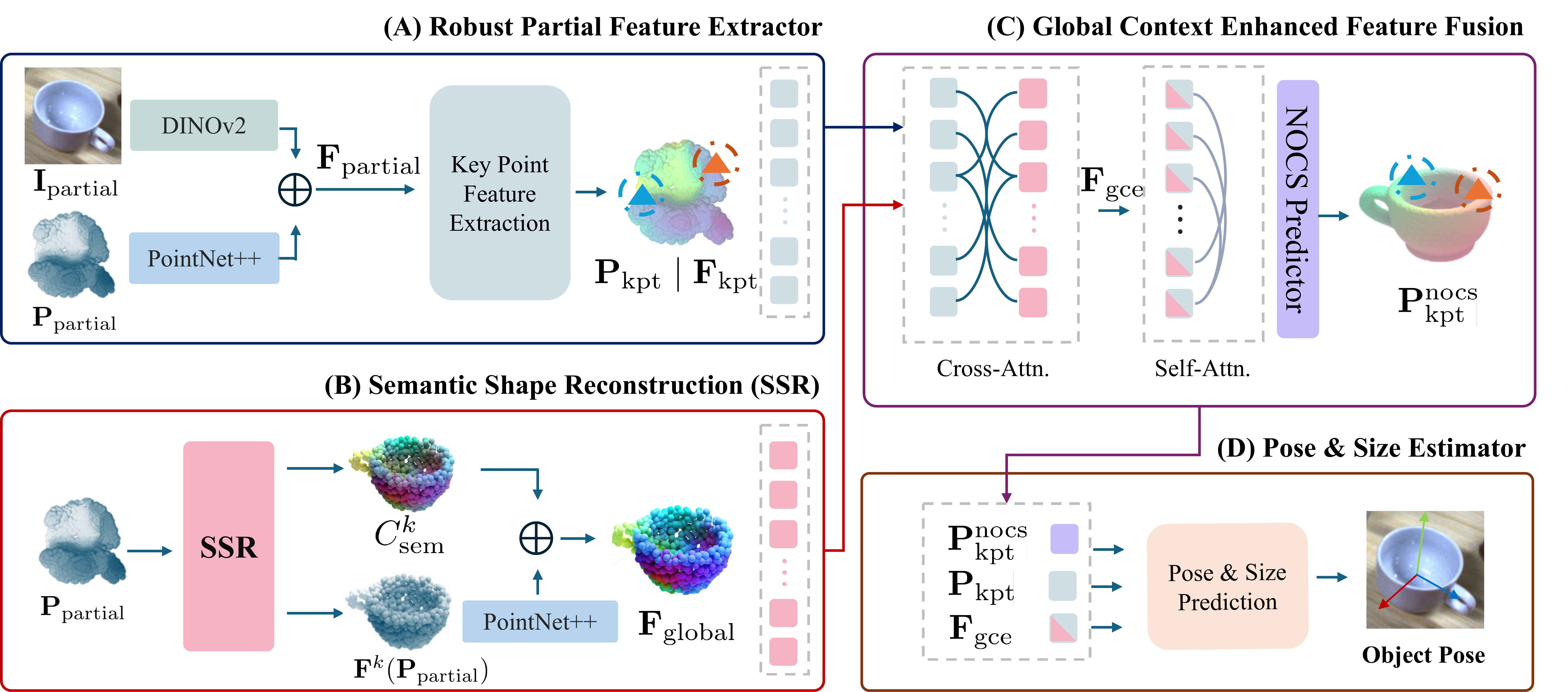}
    \vspace{-0.7cm}
    \caption{Illustration of GCE-Pose: (A) Semantic and geometric features are extracted from an RGB-D input. A keypoint feature detector identifies robust keypoints and extracts their corresponding features. (B) An instance-specific and category-level semantic global feature is reconstructed using our SSR module. (C) The global features are fused with the keypoint features to predict the keypoint NOCS coordinates. (D) The predicted keypoints, NOCS coordinates, and fused keypoint features are utilized for pose and size estimation.}
    \vspace{-0.3cm}
    \label{fig:pipeline_overview}
\end{figure*}

\section{Related Works} \label{sec:related_works}
\subsection{Object Reconstruction for Pose Estimation}
Object reconstruction is essential for object pose estimation when CAD models are unavailable, as it captures object geometry and appearance while establishing a canonical space. Methods like OnePose~\cite{sun2022onepose}, OnePose++~\cite{he2022oneposeplusplus}, and CosyPose~\cite{labbe2020cosypose} employ Structure from Motion (SfM) to match features across views, while approaches such as NeRFPose~\cite{li2023nerf}, GS-Pose~\cite{cai2024gs}, and FoundationPose~\cite{wen2024foundationpose} utilize Neural Radiance Fields or 3D Gaussian Splatting~\cite{kerbl20233dgaussian} for flexible reconstruction.

For known object categories, semantic information can enhance instance-level reconstructions. Some methods build semantic representations directly: Goodwin~\textit{et al.}~\cite{goodwin2022zero} align 3D views to a query view, and Zero123-6D~\cite{di2024zero123} synthesizes views via diffusion models to reduce reference views. I2cNet~\cite{i2c} expands such techniques to categories by integrating a 3D mesh reconstruction module. Other methods use shape priors, such as SPD~\cite{tian2020shape}, RePoNet~\cite{fu2022category}, and Wang~\textit{et al.}~\cite{wang2021category}, which learn instance reconstructions from category-specific priors. SGPA~\cite{chen2021sgpa} and RBP~\cite{zhang2022rbp} dynamically adapt priors based on observed structures, while SAR-Net~\cite{sar} and ACR-Pose~\cite{fan2021acrpose} further incorporate geometric and adversarial strategies. GS-Pose~\cite{wang2025gs} projects DINOv2~\cite{oquab2023dinov2} features onto a 3D reference shape, aiding feature alignment and pose prediction. From here, we propose a novel method that further improves the aforementioned methods by integrating semantics to the reconstructions to provide global contextual information for pose estimation.

\subsection{Representation Learning for Pose Estimation}
Learning effective feature representations from input modalities is crucial to pose estimation, evolving alongside advancements in vision neural networks. Early visual feature extractors relied on CNN backbones~\cite{xiang2017posecnn,kehl2017ssd,sundermeyer2018implicit,peng2019pvnet,zakharov2019dpod,labbe2022megapose} to predict or refine object poses from single RGB images. Recently, foundational models like DINOv2 have been widely adopted~\cite{nguyen2024gigapose,ornek2025foundpose,ausserlechner2024zs6d,chen2024secondpose,lin2024sam} to enhance robustness and contextual understanding.

Beyond 2D-only approaches, many methods now combine 2D image and 3D point cloud networks to jointly extract semantic and geometric features, constructing robust embeddings for tasks such as direct pose regression~\cite{wang2019densefusion,he2021ffb6d,chen2024secondpose,he2020pvn3d} or feature matching~\cite{huang2024matchu,caraffa2025freeze,lin2024sam}. RGB-D methods address feature fusion at multiple levels, such as 2D-3D and local-global fusion. Methods like DenseFusion~\cite{wang2019densefusion} and PVN3D~\cite{he2020pvn3d} concatenate per-pixel geometric and RGB features, while SecondPose~\cite{chen2024secondpose} employs MLPs to fuse DINOv2~\cite{oquab2023dinov2} features with Point Pair Features (PPF)~\cite{drostppf}.

More recent transformer-based approaches, including SAM6D~\cite{lin2024sam} and MatchU~\cite{huang2024matchu}, integrate both local RGB-D and global CAD model features~\cite{qin2023geotransformer,yu2023rotation,yu2024riga}, demonstrating the value of cross-modality fusion. In this work, we not only incorporate categorical semantic priors into object reconstruction but also effectively integrate global context into local embeddings through our fusion module.

Occlusions and object symmetries introduce visual ambiguities~\cite{manhardt2019explaining}, necessitating the consideration of multiple correct poses. To address this, various methods frame pose prediction as distribution estimation, learning ambiguity-aware representations~\cite{haugaard2023spyropose,vutukur2025alignist,stablepose}. We tackle the occlusion challenge by employing a completion task that enables the network to reason about full geometric representations despite missing points.

\subsection{Generalizing Object Pose Estimators} 
The constraints of 3D model-based pose estimation have been relaxed to category-level estimation, where the task is to predict the pose of an unknown instance within a known category (e.g., a fork in "cutlery") on benchmarks like NOCS~\cite{nocs}, PhoCal~\cite{wang2022phocal}, and HouseCat6D~\cite{jung2024housecat6d}. 

Category-level pose estimation aims to predict 9DoF poses for novel instances within specified categories. Wang~\textit{et al.}~\cite{nocs} introduced the Normalized Object Coordinate Space (NOCS) framework, mapping observed point clouds to a canonical space with pose recovery via the Umeyama algorithm~\cite{umeyama1991least}. Subsequent methods improve accuracy~\cite{chen2020learning,Chen_2021_CVPR,DualPoseNet,di2022gpv,Zheng_2023_CVPR,lin2024instance, zhang2024generative}. Some works adopt prior-free methods, such as VI-Net~\cite{lin2023vi}, which separates rotation components, and IST-Net~\cite{liu2023istnet}, which transforms camera-space features implicitly. AG-Pose~\cite{lin2024instance} achieves state-of-the-art results by learning keypoints from RGB-D without priors. In contrast, we incorporate learned priors for geometry and semantics to complete partial observations and map mean shape semantics onto observed instances.

Self-supervised approaches are also popular in category-level estimation, refining models without annotated real data. CPS++~\cite{manhardt2020cpspp} uses a differentiable renderer to adapt synthetic models with real, unlabeled RGB-D inputs, while other works~\cite{wang2022crocps,wang2024improving} tackle photometric challenges using RGBP polarization~\cite{gao2022polarimetric,ruhkamp2024s} and contextual language cues~\cite{wang2024improving}. Self-DPDN~\cite{lin2022category} employs a shape deformation network for self-supervision, while our approach leverages a categorical shape prior without network refinement.

In open-vocabulary settings, POPE~\cite{fan2024pope} introduces promptable object pose, (H)Oryon~\cite{corsetti2024open,corsetti2024high} use vision-language models and stereo matching, while NOPE~\cite{nguyen2024nope} and SpaRP~\cite{xu2025sparp} predict pose distributions or relative NOCS-maps. These methods often treat pose estimation as correspondence matching~\cite{corsetti2024open,fan2024pope} or reconstruction~\cite{nguyen2024nope}. We instead deform a mean shape within an absolute category space to capture instance-specific correspondences.

\section{Method} \label{sec:method}
The objective of GCE-Pose is to estimate the 6D object pose and size from RGB-D data. Given a single RGB-D frame and the category instance mask, we obtain the partial RGB observation $\mathbf{I}_{\text{partial}}$ and its corresponding partial point cloud $\mathbf{P}_{\text{partial}}$ derived from the depth map. Utilizing $\mathbf{I}_{\text{partial}}$ and $\mathbf{P}_{\text{partial}}$, the objective is to recover the 3D rotation \( \mathbf{R} \in \text{SO}(3) \), the 3D translation \( \mathbf{t} \in \mathbb{R}^3 \), and the size \( \mathbf{s} \in \mathbb{R}^3 \) of the target object.

GCE-Pose consists of four main modules (\cref{fig:pipeline_overview}): Robust Partial Feature Extraction (\cref{subsec:keypoint_feature}), Semantic Shape Reconstruction (\cref{subsec:dlssr}), Global Context Enhanced Feature Fusion (\cref{subsec:gce}), and Pose \& Size Estimator (\cref{subsec:pse}).

\subsection{Robust Partial Feature Extraction}
\label{subsec:keypoint_feature}
Partial observations from RGB-D sensors often contain significant noise and incomplete geometry, making dense correspondence prediction unreliable. We address this challenge with a keypoint-based approach~\cite{lin2024instance} that focuses on the most discriminative and reliable object regions.

The $N$ input points are put in order within $\mathbf{P}_{\text{partial}} \in \mathbb{R}^{N \times 3}$ and we extract point features $\mathbf{F}_{\text{P}} \in \mathbb{R}^{N \times C_1}$ using PointNet++ \cite{qi2017pointnetplusplus}. For the RGB image $\mathbf{I}_\text{\text{partial}}$, we extract the image feature $\mathbf{F}_{\text{I}} \in \mathbb{R}^{N \times C_2}$ using DINOv2~\cite{oquab2023dinov2} and concatenate $\mathbf{F}_{\text{I}}$ to $\mathbf{F}_{\text{P}}$ to obtain $\mathbf{F}_{\text{partial}} \in \mathbb{R}^{N \times C}$.
We follow AG-Pose~\cite{lin2024instance} for keypoint detection. First, $M$ keypoint features are extracted using a learnable embedding $\mathbf{F}_{\text{emb}} \in \mathbb{R}^{M \times C}$, which undergoes cross-attention with $\mathbf{F}_{\text {partial}}$ to attend to critical regions in $\mathbf{P}_{\text{partial}}$. This process yields a feature query matrix $\mathbf{F}_{\mathrm{q}} = \text{CrossAttention}(\mathbf{F}_{\text{emb}}, \mathbf{F}_{\text{partial}})$. We then compute correspondences via cosine similarity, forming a matrix $\mathbf{A} \in \mathbb{R}^{M \times N}$, and select $M$ keypoints from $\mathbf{P}_{\text{partial}}$ as $\mathbf{P}_{\mathrm{kpt}} = \operatorname{softmax}(\mathbf{A}) \mathbf{P}_{\text{partial}}$.
To ensure keypoints lie on the object surface and minimize outliers, an object-aware Chamfer distance loss \(\mathcal{L}_{\text{ocd}}\) is applied. With ground truth pose $\mathbf{T}_{\text{gt}}$, we filter outliers by comparing each point $x \in \mathbf{P}_{\text{partial}}$ to the instance model $\mathbf{M}_{\text{obj}}$:
\begin{equation}
\vspace{-0.2cm}
\min_{y \in \mathbf{M}_{\text{obj}}} \left\| \mathbf{T}_{\text{gt}}(x) - y \right\|_2 < {\tau_1},
\label{equ:outlier_threshold}
\end{equation}
where ${\tau_1}$ is an outlier threshold. The object-aware Chamfer distance loss is then:
\begin{equation}
\vspace{-0.2cm}
\mathcal{L}_{\text{ocd}} = \frac{1}{|\mathbf{P}_{\text{kpt}}|} \sum_{x \in \mathbf{P}_{\text{kpt}}} \min_{{y} \in \mathbf{P}_{\text{partial}}^*} \|{x} - {y}\|_2.
\vspace{-0.1cm}
\label{equ:ocd}
\end{equation}
To prevent keypoints from clustering, a diversity regularization loss is added:
\begin{equation}
\vspace{-0.1cm}
\mathcal{L}_{\text{div}} = \sum_{x \neq y \in \mathbf{P}_{\text{kpt}}} \max \{ 0, {\tau}_2 - \|x - y\|_2 \},
\vspace{-0.1cm}
\label{equ:div}
\end{equation}
where ${\tau}_2$ controls keypoint distribution.
To enhance features with geometric context, the Geometric-Aware Feature Aggregation (GAFA) module~\cite{lin2024instance} is applied. GAFA augments each keypoint with (1) local geometric details from K-nearest neighbors and (2) global information from all keypoints, improving feature discriminability for correspondence estimation.

\begin{figure*}[t]
    \centering
\includegraphics[width=\textwidth]{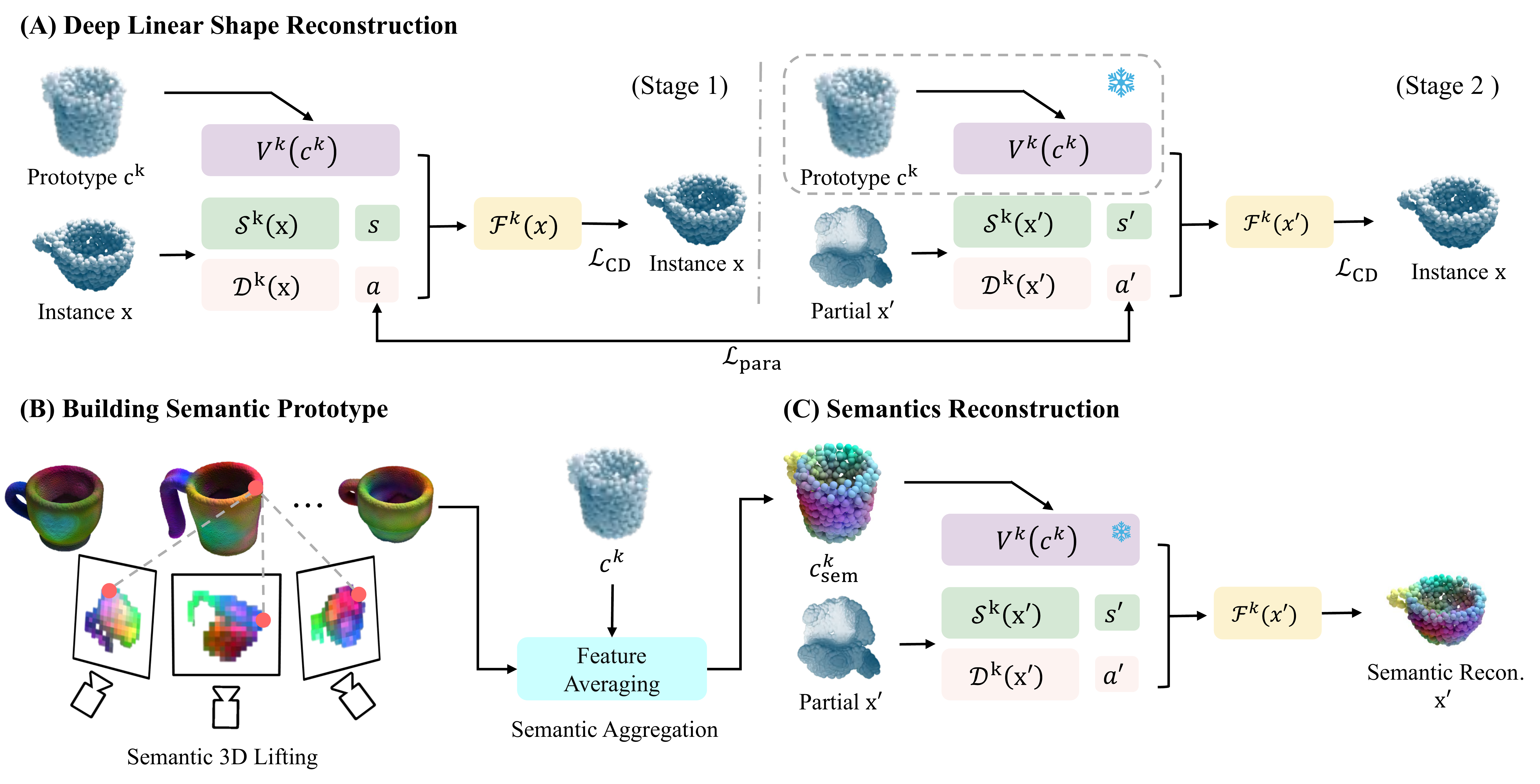}
    \vspace{-0.6cm}
    \caption{Illustration of Deep Linear Semantic Shape Model. A Deep Linear Semantic Shape model is composed of a prototype shape $c$, a scale network $\mathcal{S}$, a deformation network $\mathcal{D}$, a Deformation field $\mathcal{V}$ and a category-level semantic features $c^k_\text{sem}$. At stage 1, we build a Deep Linear Shape (DLS) model using sampled point clouds from all ground truth instances within each category, training a linear parameterization network to represent each instance. At stage 2, we retrain the DLS model to regress the corresponding DLS parameters from partial point cloud inputs using a deformation and scale network. During testing, the network predicts DLS parameters for unseen objects and reconstructs their point clouds based on the learned deformation field to get semantic reconstruction.}
    \vspace{-0.3cm}
    \label{fig:pipeline_deep_lineaer}
\end{figure*}

\subsection{Semantic Shape Reconstruction} \label{subsec:dlssr}
Intra-class variation is a key challenge in category-level pose estimation. To tackle this issue, category-level shape priors have been extensively used in object pose estimation. By representing the shape with mean shapes and deformations~\cite{tian2020shape} or learning implicit neural representations for geometry recovery~\cite{irshad2022shapo, irshad2022centersnap}, pose estimators can better learn correspondences in NOCS space, benefiting from accurate shape priors. While geometric shape reconstruction provides valuable priors, it cannot fully capture the rich semantic information of object parts. Recent advances in 2D foundation models, particularly DINO~\cite{oquab2023dinov2}, have demonstrated remarkable capabilities in extracting zero-shot semantic information from single RGB images. Building upon this insight, we propose \textbf{Semantic Shape Reconstruction (SSR)} to learn a per-category linear shape model similar to \cite{loiseau2021representing} that describes an object using instance-specific geometry and category-level semantic features.

\noindent\textbf{Deep Linear Semantic Shape Model.}  
To overcome the challenges posed by partial observations from depth sensors, such as occlusions and incomplete geometry, we employ a variation of the deep linear shape model \cite{loiseau2021representing}. This approach is motivated by the need to robustly and efficiently parameterize object shapes with shape parameters and produce a completed 3D object representation, even when faced with limited input data. We represent each point in our model as a tuple $(x, f)$ where $x \in \mathbb{R}^3$ represents a spatial coordinate and $f \in \mathbb{R}^C$ represents its semantic feature vector. For $I$ points of an object instance within category $k$, we learn a linear shape model. The model for category $k$ consists of (i) a geometric prototype ${c}^k \in \mathbb{R}^{I \times 3}$ with associated semantic features ${c}^{k}_{\text{sem}} \in \mathbb{R}^{I \times C}$, (ii) a set of geometric deformation basis vectors ${v}^k = \{{v}_1^k, \ldots, v_D^k\}$ where ${v}_i^k \in \mathbb{R}^{I \times 3}$, and (iii) a scale parameter vector ${s^k} \in \mathbb{R}^3$.
The key insight of our approach is that semantic features remain coupled to their corresponding points during geometric deformation. Any semantic shape $\mathbf{U}_{\text{k}}$ in the model family is defined by:
\begin{equation}
\vspace{-0.1cm}
\mathbf{U}_{\text{k}} = (\mathbf{X}_{\text{k}}, \mathbf{F}_{\text{k}})
= \left( s^k \odot (c^k + \sum_{i=1}^D a^k_i v^k_{i}), c^k_{\text{sem}} \right)
\vspace{-0.0cm}
\label{equ:dls_familiy}
\end{equation}
where $\mathbf{X}_{\text{k}} \in \mathbb{R}^{I \times 3}$ are the $I$ points in shape prior $k$ and $\mathbf{F}_{\text{k}} \in \mathbb{R}^{I \times C}$ their associated features.
The shape parameter vector is given by ${a^k} = \left(a^k_1, \ldots, a^k_D\right) \in \mathbb{R}^D$, ${s^k} \in \mathbb{R}^3$ controls scaling, and $\odot$ defines the element-wise Hadamard product.
We train two neural networks for each category $k$ to predict shape parameter $a^k$ with network $\mathcal{D}^k$ and scale $s^k$ with $\mathcal{S}^k$.

To optimize the model, we minimize the Chamfer distance loss, $\mathcal{L_{\text{CD}}}$, which ensures accurate shape reconstruction through:
\begin{equation}
\vspace{-0.1cm}
\mathcal{L_{\text{CD}}} = \sum_{x \in \mathbf{P}} \min_{k} d\left(x, \mathbf{U}_{\text{k}}\right),
\label{equ:l_cd}
\vspace{-0.1cm}
\end{equation}
with the Chamfer distance $d$, ground truth point clouds $\mathbf{P}$ from category $k$ and shape reconstruction $\mathbf{U}_{\text{k}}$ defined in \cref{equ:dls_familiy} . Training with ground truth yields the optimal parameters $\bar{a}^k$, $\bar{s}^k$, $c^k$, and $v^k$ which allow to formulate an additional loss to refine shape reconstruction under partial observations $\mathbf{P}_{\text{partial}}$ by freezing $c^k$, and $v^k$ within
\begin{equation}
\vspace{-0.1cm}
\small
\mathcal{L}_{\text {para}} = \sum_{x' \in \mathbf{P}_{\text{partial}}} \lambda_1 \left| \mathcal{D}^k(x') - \bar{a}^k \right| + \lambda_2 \left| \mathcal{S}^k(x') - \bar{s}^k \right|.
\label{equ:l_para}
\vspace{-0.1cm}
\end{equation}
Finally, we combine the reconstruction and parameter loss to formulate the overall loss
\begin{equation}
\mathcal{L}_{\text{rec}} = \lambda_{\text{CD}} \cdot \mathcal{L}_{\text{CD}} + \lambda_{\text{para}} \cdot \mathcal{L}_{\text{para}},
\label{equ:l_rec}
\end{equation}
where $\lambda_{\mathrm{CD}}$ and $\lambda_{\text {para }}$ are the hyperparameters that weight the contributions of $\mathcal{L}_{\mathrm{CD}}$ and $\mathcal{L}_{\text {para }}$, respectively.

\noindent{\textbf{Semantic Prototype Construction.}}
To effectively integrate rich semantic information into our 3D shape reconstruction, we employ a process that begins by extracting dense semantic features from multiple RGB images of each object instance using the DINOv2 ~\cite{oquab2023dinov2}. For each object instance without texture, we position multiple virtual cameras around the object to capture RGB images and depth maps from diverse viewpoints. This setup ensures full coverage of the object's surface and mitigates occlusion effects. The RGB images are processed through the DINOv2 model to extract dense 2D semantic feature maps.  Using the corresponding depth maps and known camera intrinsics and extrinsic, we project the 2D semantic features into 3D space. For each pixel $(u, v)$ in the image, we compute its 3D position ${P}$ using the depth value $z$ and project the associated semantic feature $\mathbf{f}_{2 \mathrm{D}}(u, v)$ to this point ${P}=z K^{-1}[u, v,1]^T,$, where $K$ is the camera intrinsic matrix. As a result, we obtain a dense semantic point cloud $\mathbf{F}_{\text{sem}}$. To ensure computational efficiency and point-wise correspondence, we downsample this dense semantic point cloud to $I$ points aligned with our geometric reconstruction. For each point $P_i$ in the deep linear shape reconstruction, we aggregate semantic features from its k nearest neighbors in the dense cloud:
\begin{equation}
\vspace{-0.1cm}
\mathbf{F}_{\text{instance}}(P_i) = \frac{1}{k}\sum_{P_j \in N_k(P_i)} \mathbf{F}_{\text{sem}}(P_j).
\vspace{-0.1cm}
\end{equation}
The category-level semantic prototype $c^k_{\text{sem}}$ is then constructed by averaging $N$ instance features across the category $k$ while maintaining point-wise correspondence with the geometric prototype $c^k$:
\begin{equation}
\vspace{-0.1cm}
c^k_{\text{sem}} = \frac{1}{N}\sum_{i=1}^N \mathbf{F}_{\text{instance}}(P^k_i)
\vspace{-0.1cm}
\label{equ:knn}
\end{equation}

\noindent{\textbf{Semantic Reconstruction.}}
The key advantage of our approach is that semantic reconstruction becomes straightforward once the semantic prototype is established. Given a partial point cloud $x'$, we first reconstruct its geometry and then directly inherit the semantic feature from the prototype.

\subsection{Global Context Enhanced Feature Fusion} \label{subsec:gce}
Traditional pose estimation methods rely primarily on partial observations, but they often struggle with challenges such as occlusion and viewpoint variations. To overcome these limitations, we propose a \textbf{Global Context Enhanced (GCE)  Feature Fusion} module that effectively integrates complete semantic shape reconstructions with partial observations, establishing robust feature correspondences.

Firstly, we aim to extract global features from our semantic reconstruction. Given partial observation $\mathbf{P}_{\text{partial}}$, we can reconstruct the shape \( \mathbf{P}_{\text{global}}\) using \cref{equ:dls_familiy}. We leverage PointNet++~\cite{qi2017pointnetplusplus} to obtain a geometry feature from $ \mathbf{P}_{\text{global}}$, then concatenate it with category-level semantic feature $c_{\text{sem}}$ to obtain the global feature $\mathbf{F}_{\text{global}} \in \mathbb{R}^{I \times C}$.

Given keypoint Features \( \mathbf{F}_{\text{kpt}} \in \mathbb{R}^{M \times C} \) and global feature \(\mathbf{F}_{\text{global}} \in \mathbb{R}^{I \times C} \), our goal is to enrich keypoint features with global semantic context. The primary challenge lies in bridging the domain gap between partial observations, captured in noisy camera-space coordinates, and the global reconstruction, represented in normalized object-space coordinates with complete shape information.

To fuse the partial feature $\mathbf{F}_{\text{kpt}}$ with $\mathbf{F}_{\text{global}}$, we transform both the partial and global features by concatenating learnable positional embedding network that maps these 3D  positions \( \mathbf{P}_{\text{kpt}}\) and \( \mathbf{P}_{\text{global}}\). into high-dimensional positional tokens, respectively. 
\begin{equation}
\small
\mathbf{F}_{\text{kpt}}' = \text{concat}(\mathbf{F}_{\text{kpt}}, \text{PE}_{\text{kpt}}), \quad \mathbf{F}_{\text{global}}' = \text{concat}(\mathbf{F}_{\text{global}}, \text{PE}_{\text{global}})
\end{equation}
where \text{PE} is positional encoding network.
We then apply an attention mechanism to merge both sources of information, where the global features provide the semantic context for refining the keypoint features.
We first project keypoint features and global features into a shared embedding space:
\begin{equation}
    \mathbf{F}_{\text{kpt}}'' = \text{LayerNorm}(\text{MLP}_{\text{proj}}(\mathbf{F}_{\text{kpt}}'))
\end{equation}
\begin{equation}
    \mathbf{F}_{\text{global}}'' = \text{LayerNorm}(\text{MLP}_{\text{proj}}(\mathbf{F}_{\text{global}}'))
\end{equation}
then, the global context enhancement is aggregated through cross-attention and residual connection:
\begin{equation}
    \mathbf{F}_{\text{context}} = \text{CrossAttn}(\mathbf{F}_{\text{kpt}}'', \mathbf{F}_{\text{global}}'')
\end{equation}
\begin{equation}
    \mathbf{F}_{\text{gce}} = \mathbf{F}_{\text{kpt}} + \mathbf{F}_{\text{context}}
\end{equation}
After fusing the keypoint $\mathbf{F}_{\text{kpt}}$ feature with our global feature $\mathbf{F}_{\text{global}}$. The resulting global context enhanced keypoint feature $\mathbf{F}_{\text{gce}}$ are then passed through the self-attention module and MLP following \cite{lin2022category} to predict the corresponding NOCS coordinates $\mathbf{P}_{\text{kpt}}^{\text{nocs}}$.

To ensure that our keypoints and associated features effectively represent the partial observation \( \mathbf{P}_{\text{partial}} \), we additionally employ a reconstruction module to recover its 3D geometry. This module takes keypoint positions and features as input, applies positional encoding to the keypoints, and refines their features through a MLP. The encoded and refined features are aggregated, and a shape decoder predicts reconstruction deltas to recover the geometry. The reconstruction loss is defined as the object-aware Chamfer distance (CD) between the partial observation \( \mathbf{P}_{\text{partial}} \) and the reconstructed point cloud \( \mathbf{P}_{\text{recon}} \) following \cref{equ:outlier_threshold}:
\begin{equation}
\mathcal{L}_{\text{rec}}=\frac{1}{\left|\mathbf{P}_{\text{recon}}\right|} \sum_{x \in \mathbf{P}_{\text{recon}}} \min _{y \in \mathbf{P}_{\text{partial}}^{\star}}\left\|x-y\right\|_2 .
\label{equ:loss_rec}
\end{equation}

\subsection{Pose Size Estimator} \label{subsec:pse}
Given the NOCS coordinates of keypoints, \(\mathbf{P}_{\text{kpt}}^{\text{nocs}} \in \mathbb{R}^{M \times 3}\), the enhanced keypoint features \(\mathbf{F}_{\text{gce}}\) and the position of keypoint $\mathbf{P}_{\text{kpt}}$, we can establish keypoint-level correspondences, which are then used to regress the final pose and size parameters, \(\mathbf{R}\), \(\mathbf{t}\), and \(\mathbf{s}\). The process is formulated as follows:
\begin{equation}
\vspace{-0.1cm}
\mathbf{f}_{\text{pose}} = \operatorname{concat} \left[ \mathbf{P}_{\text{kpt}}, \mathbf{F}_{\text{gce}}, \mathbf{P}_{\text{kpt}}^{\text{nocs}} \right]
\end{equation}
\begin{equation}
\small
\left( \mathbf{R}, \mathbf{t}, \mathbf{s} \right) =
\left( \operatorname{MLP}_R \left( \mathbf{f}_{\text{pose}} \right), \operatorname{MLP}_t \left( \mathbf{f}_{\text{pose}} \right), \operatorname{MLP}_s \left( \mathbf{f}_{\text{pose}} \right) \right)
\end{equation}

For the rotation representation \(\mathbf{R}\), we utilize the 6D continuous representation proposed in \cite{Zhou_2019_CVPR}. For the translation \(\mathbf{t}\), we adopt the strategy from \cite{Zheng_2023_CVPR} by predicting the residual translation between the ground truth and the mean position of the point cloud.

\begin{figure*}[t]
    \centering
    \includegraphics[width=0.95\textwidth]{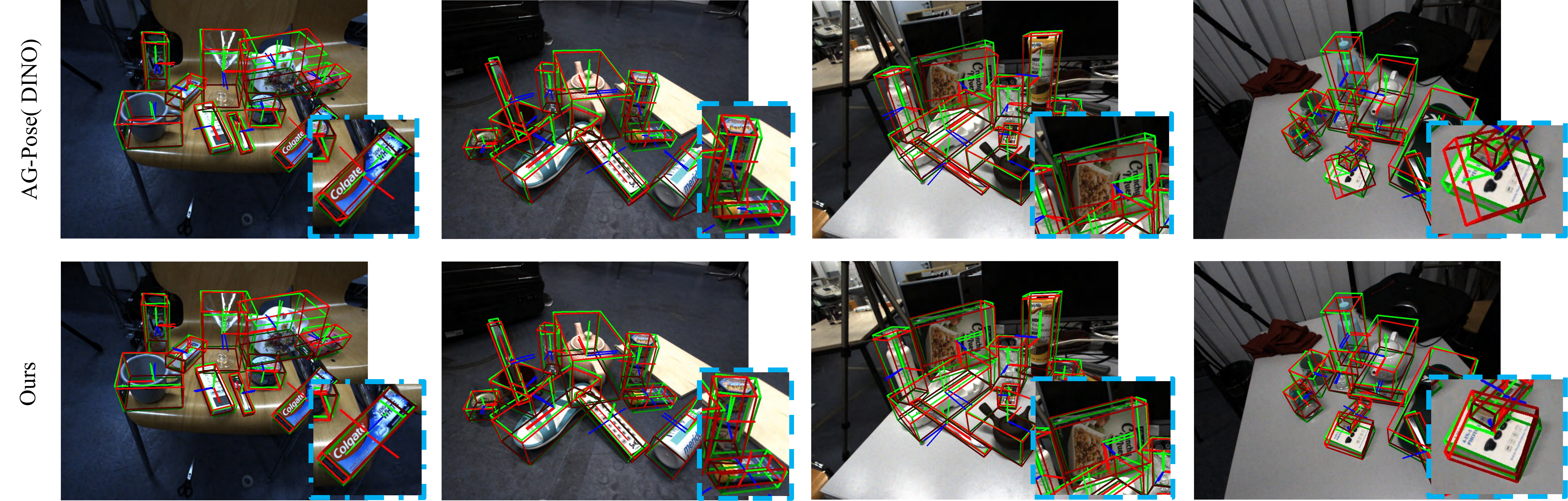}
    \vspace{-0.2cm}
    \caption{Visualization of category-level object pose estimation results on HouseCat6D dataset~\cite{jung2024housecat6d}. Predicted 3D bounding boxes are shown in red, with ground truth in green. Challenging cases are highlighted in pink side squares. Leveraging our global context-enhanced pose prediction pipeline, GCE-Pose outperforms the SOTA AG-Pose~\cite{lin2024instance} (DINO), demonstrating robustness to occlusions and strong generalization to novel instances.
}
    \vspace{-0.2cm}
    \label{fig:vis_housecat_pred}   
\end{figure*}

\begin{table*}[h]
\centering
\resizebox{\textwidth}{!}{%
\begin{tabular}{l|l|c c|c c c c | c c} 
\hline
\hline
\textbf{Dataset} & \textbf{Method} & \textbf{Shape Prior} & \textbf{Semantic Prior} & \textbf{5°2cm} & \textbf{5°5cm} & \textbf{10°2cm} & \textbf{10°5cm}  & \textbf{IoU50} & \textbf{IoU75}\\
\hline

\multirow{4}{*}{\centering HouseCat6D~\cite{jung2024housecat6d}} 
 & VI-Net~\cite{lin2023vi}     &  &  & 8.4  & 10.3 & 20.5 & 29.1 & 56.4 & - \\
 & SecondPose~\cite{chen2024secondpose} &  &   & 11.0 & 13.4 & 25.3 & 35.7  & 66.1 & - \\
 & AG-Pose~\cite{lin2024instance} &  &  & 11.5 & 12.0 & 32.7 & 35.8 & 66.0 & 45.0   \\
 & AG-Pose (DINO)~\cite{lin2024instance}  &  &  & 21.3 & 22.1 & 51.3 & 54.3 & 76.9 & 53.0  \\

 & \textbf{GCE-Pose} (Ours) & \(\checkmark\) & \(\checkmark\)  
 & \textbf{24.8} & \textbf{25.7} & \textbf{55.4} & \textbf{58.4} & \textbf{79.2} & \textbf{60.6} \\
\hline
\hline
\multirow{8}{*}{\centering NOCS-REAL275~\cite{nocs}}   

& SecondPose~\cite{chen2024secondpose}     &  &        & 56.2 & 63.6 & 74.7 & 86.0  & -    & -   \\
& AG-Pose~\cite{lin2024instance}      &  &     & 56.2 & 62.3 & 73.4 & 81.2  & 83.8 & 77.6 \\
& AG-Pose(DINO)~\cite{lin2024instance}    &  &     & \textbf{57.0} & 64.6 & 75.1 & 84.7  & \textbf{84.1} & \textbf{80.1}\\
\cdashline{2-10} 

& RBP-Pose~\cite{zhang2022rbp} & \(\checkmark\) &   & 38.2 & 48.1 & 63.1 & 79.2   & -    &   67.8   \\
& DPDN~\cite{lin2022category} & \(\checkmark\) &    & 46.0 & 50.7 &  70.4    & 78.4    & 83.4    &  76.0 \\
& Query6DoF~\cite{query6dof}  & \(\checkmark\) &   & 49.0   & 58.9 & 68.7 & 83.0 & 82.5 & 76.1   \\
& GS-Pose~\cite{wang2025gs}         & \(\checkmark\) & \(\checkmark\)  & -    & 28.8 & -    & 60.1   & -    & 63.2  \\
&\textbf{GCE-Pose} (Ours)  & \(\checkmark\) & \(\checkmark\) & \textbf{57.0} & \textbf{65.1} & \textbf{75.6} &
\textbf{86.3}  & \textbf{84.1} & 79.8\\
\hline
\hline
\end{tabular}%
}
\vspace{-0.2cm}
\caption{Quantitative comparison of category-level object pose estimation on HouseCat6D and NOCS-REAL275 datasets.}
\vspace{-0.5cm}
\label{tab:merged_results}
\end{table*}

\subsection{Overall Loss Function} \label{subsec:pse}
The overall loss function for pose estimation is as follows:
\begin{equation}
\mathcal{L}_{\text{all}} = \lambda_1 \mathcal{L}_{\text{ocd}} + \lambda_2 \mathcal{L}_{\text{div}} + \lambda_3 \mathcal{L}_{\text{rec}} + \lambda_4 \mathcal{L}_{\text{nocs}} + \lambda_5 \mathcal{L}_{\text{pose}}
\label{equ:overall_loss}
\end{equation}
where $\lambda_1, \lambda_2, \lambda_3, \lambda_4, \lambda_5$ are hyperparameters to balance the contribution of each term. For $\mathcal{L}_{\text{pose}}$, we use:
\begin{equation}
\mathcal{L}_{\text{pose}} = \|\mathbf{R}_{\text{gt}} - \mathbf{R}\|_{\rm F} + \|\mathbf{t}_{\text{gt}} - \mathbf{t}\|_2 + \|\mathbf{s}_{\text{gt}} - \mathbf{s}\|_2.
\end{equation}

We generate ground truth NOCS coordinates of keypoints $\mathbf{P}_{\text{kpt}}^{\text{gt}}$ by projecting their coordinates under camera space $\mathbf{P}_{\text{kpt}}$ into NOCS using the ground-truth \(\mathbf{T}_{\text{gt}} = \left( \mathbf{R}_{\text{gt}}, \mathbf{t}_{\text{gt}}, \mathbf{s}_{\text{gt}} \right)\). For $\mathcal{L}_{\text{nocs}}$, we use the Smooth $L_1$ loss with
\begin{equation}
\mathcal{L}_{\text{nocs}} = \| \mathbf{T}_{\text{gt}}(\mathbf{P}_{\text{kpt}}^{\text{gt}}) - \mathbf{P}_{\text{kpt}}^{\text{nocs}} \|_{\text{SL1}}
\end{equation}

\section{Experiment} \label{sec:experiment}
\begin{figure*}[t] 
    \hspace*{\fill} 
    \begin{minipage}[t]{1.0\textwidth} 
        \centering
        \includegraphics[width=0.95\textwidth]{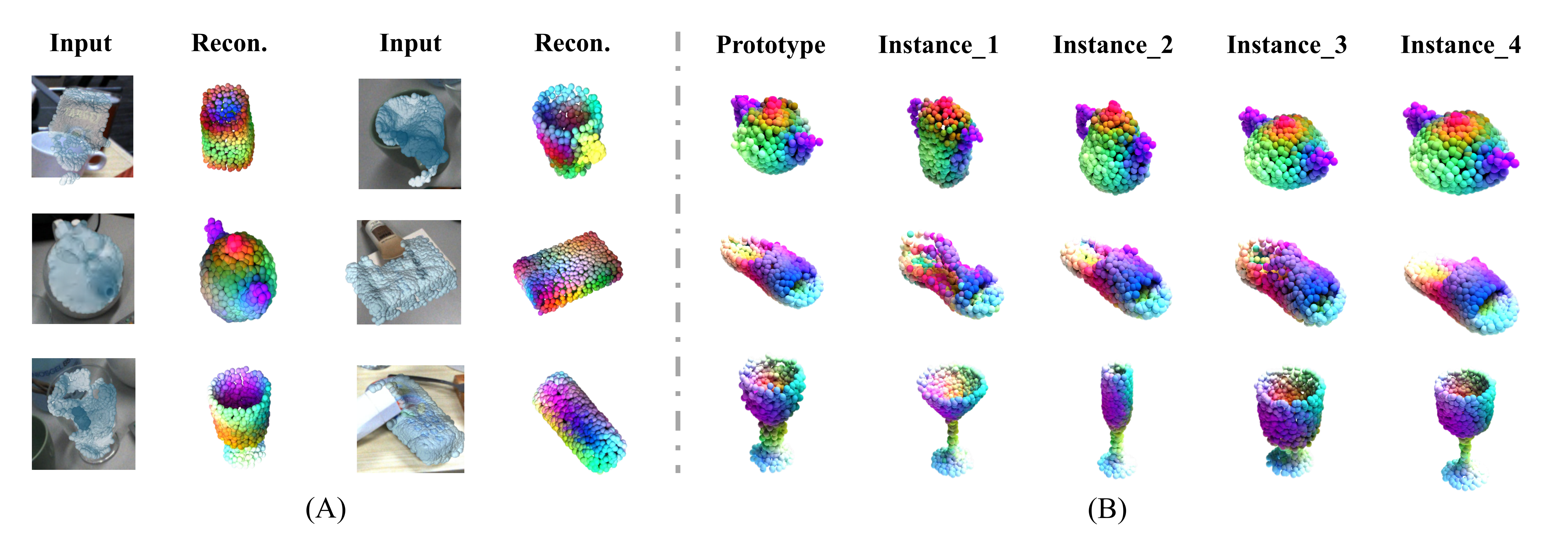} 
        \vspace{-0.5cm}
        \caption{(A) shows the visualization of input partial points and the output semantic shape reconstructions; (B) visualizes the semantic prototypes of different categories and the aggregated instance semantics.}
        \vspace{-0.4cm}
        \label{fig:vis_intra_variance}
    \end{minipage}
\end{figure*}

\subsection{Implementation Details}
For the HouseCat6D dataset \cite{jung2024housecat6d}, cropped images are resized to 224 $\times$ 224 for feature extraction, and 1024 points are sampled from inputs. For Partial Feature Extraction, the number of keypoints is $M = 96$, and the feature dimensions for geometric and DINO features are $C_1 = 128$, $C_2 = 128$, and $C = 256$. In deep linear shape reconstruction, we set the basis dimension $D$ to 5, and the number of points in the prototype is 1024. The pose estimation network is trained with batch size 36 with an ADAM \cite{Kingma2014AdamAM} optimizer with a triangular2 cyclical learning rate schedule \cite{smith2017cyclical} on a single NVIDIA 4090 GPU for 150 epochs. We attach more implementation details in our Appendix.

\subsection{Evaluation Benchmarks} 
\noindent{\textbf{Datasets.}} We evaluate our method on two challenging real-world benchmarks: HouseCat6D~\cite{jung2024housecat6d} and NOCS-REAL275~\cite{nocs}. HouseCat6D contains 21K images of 194 household instances across 10 categories, with 2,929 images of 50 instances reserved for evaluation, covering diverse shapes, occlusions, and lighting. NOCS-REAL275 consists of 7K images across 6 object categories and 13 scenes, with 4.3K images used for training and 2,750 for testing. We compare our method against baselines on both datasets with the same setup for segmentation mask and conduct the ablation studies on HouseCat6D.
\noindent{\textbf{Evaluation Metrics.}} Following prior work~\cite{lin2024instance, chen2024secondpose, lin2023vi, di2022gpv}, we evaluate performance with two metrics:
{$n^{\circ} m$ cm} and \textit{3D IoU}. $n^{\circ} m$ cm metric computes mean Average Precision (mAP) for rotation and translation accuracy, considering predictions correct if the rotation error is within $n^{\circ}$ and translation error within $m$ cm.
\textit{3D IoU} is a  mAP-based metric that assesses 3D bounding box IoU with thresholds at $50\%$ and $75\%$, capturing both pose and object size.
\vspace{-0.2cm}
\subsection{Comparison with the State-of-the-Art} 
\label{subsec:comp_sota}
\noindent{\textbf{Results on HouseCat6D.}} Table~\ref{tab:merged_results} presents GCE-Pose’s performance on the HouseCat6D dataset, where it outperforms state-of-the-art methods across all metrics against the existing approaches. Compared to AG-Pose, even strengthened with DINOv2~\cite{lin2024instance}, our method shows $16\%$ improvement on the most strict 5°2cm metric and notable gains on other metrics, demonstrating the efficacy of integrating global context priors into the local feature-based pose estimation pipeline. Against SecondPose~\cite{chen2024secondpose}, which fuses PPF geometric and DINOv2 semantic features from only the partial observations, our method achieves more than $30\%$ improvement in IoU75 metric and over $100\% $ in {$n^{\circ} m$ cm} metrics. Qualitative comparisons of our method against AG-Pose are shown in Figure~\ref{fig:vis_housecat_pred}.

\noindent{\textbf{Results on NOCS-REAL275.}} Table~\ref{tab:merged_results} also shows GCE-Pose’s performance on NOCS-REAL275 dataset, achieving the highest scores on most metrics. GCE-Pose surpasses prior-free baselines by integrating robust global context priors into a strong pose estimator, proving the benefits of our design. We also compare with methods using categorical priors: RBP-Pose~\cite{zhang2022rbp}, Query6DoF~\cite{query6dof} and DPND~\cite{lin2022category}, which rely solely on shape priors, while GS-Pose~\cite{wang2025gs} uses both shape and semantic priors with a single-instance reference. In contrast, GCE-Pose learns robust deformation priors across multiple instances and completes partial inputs into a full semantic shape, outperforming all prior-based methods and by more than $23\%$ on 5°2cm metric, demonstrating the effectiveness of our SSR and GCE fusion modules.


\begin{table}[ht]
    \centering
    \vspace{-0.2cm}
    \resizebox{\linewidth}{!}{
    \begin{tabular}{l|c|c|c|c|c|c}
        \hline
       \textbf{ Method} & \textbf{Ins. recon.} & \textbf{Mean shape} & \textbf{Geo.} & \textbf{Sem.} & $\textbf{5°2cm}$ & $\textbf{5°5cm}$  \\
        \hline
        (0) AG-Pose DINO (Baseline) & \(\times\) & \(\times\)& \(\times\) & \(\times\) &  21.3 & 22.1\\
        (1) Ours (Instance Geo.) & \(\checkmark\) & \(\times\)& \(\checkmark\) & \(\times\)& 22.2    & 23.7 \\
        (2) Ours (Categorical Sem.) & \(\times\) & \(\checkmark\)& \(\times\) & \(\checkmark\)& 22.7  & 24.3  \\
        (3) Ours (Mean shape Geo. \& Categorical Sem.) & \(\times\) & \(\checkmark\)& \(\checkmark\) & \(\checkmark\)&  23.4 & 24.2 \\
        (4) GCE-Pose (Full pipeline) & \(\checkmark\) & \(\checkmark\)& \(\checkmark\) & \(\checkmark\) &  \textbf{24.8} & \textbf{25.7} \\
        \hline
    \end{tabular}
    }
    \vspace{-0.2cm}
    \caption{\footnotesize Ablation study on different global priors.
    }
    \vspace{-0.6cm}
    \label{tab:method_dcomparison}
\end{table}

\subsection{Ablation Studies}
To show the efficacy of our design choices, we conduct exhaustive ablation experiments on the HouseCat6D dataset.
\noindent\textbf{Effects of Global Context Priors.}
We evaluate the effect of global context priors on the pose estimation backbone by comparing different global prior configurations, including: 
\textit{(0) Baseline.} AG-Pose with the DINOv2 backbone, excluding SSR and GCE fusion modules (no global shape or semantic priors). 
\textit{(1) Instance geometric prior.} Geometric features alone are passed into the GCE fusion module, omitting semantic information.
\textit{(2) Categorical semantic prior.} The shape reconstruction module is excluded, and the categorical semantic prior is fused with local features.
\textit{(3) Mean shape geometric \& Categorical semantic prior.} The SSR module applies mean shape reconstruction and categorical semantic features.
\textit{(4) GCE-Pose (Full pipeline).} Our full method combines instance-specific geometric and categorical semantic priors.
As shown in Table~\ref{tab:method_dcomparison}, \textit{(1)} yields a $4\% $ improvement on the 5°2cm metric over the baseline, \textit{(2)} yields $7\%$, and \textit{(3)} achieves $10\%$. Our method (5), combining instance geometric and semantic priors as global context guidance, reaches the highest performance and surpasses the baseline by $16\%$. 

\noindent{\textbf{Robustness of Semantic Shape Reconstruction.}}
Qualitative results of our semantic shape reconstruction are shown in Figure~\ref{fig:vis_intra_variance}. Specifically, (a) shows partial input points and their reconstructed points with semantic features, illustrating the robustness of our SSR module to noisy and occluded scenarios. (b) displays the categorical semantic prototype where the 3D semantics are lifted from DINO features. We also show the instances where the semantics are aggregated from our Deep Linear Semantic Shape Models. We visualize the semantic features by sharing the PCA centers, demonstrating our semantic feature aggregation is robust against shape variance.

\noindent{\textbf{Effects of the GCE Feature Fusion Module.}}
In Table~\ref{tab:gce_fusion}, we evaluate our GCE feature fusion by experimenting with different DINOv2 tokens. Our results indicate that using the key tokens to aggregate the global semantic prior yields better pose estimation performance than using value tokens. We attribute this improvement to the design of our cross-attention layer where the local features serve as the query and global semantic prior as the key and value. This setup allows the global semantic prior to initialize the attentional weights between local and global cues effectively, enhancing the feature fusion for pose estimation.
\vspace{-0.2cm}

\begin{table}[t]
\centering
\resizebox{\columnwidth}{!}{%
\begin{tabular}{c|c c c c|c c } 
\hline
\textbf{Fusion features} &   \textbf{5°2cm} & \textbf{5°5cm} & \textbf{10°2cm} & \textbf{10°5cm} & \textbf{IoU50} & \textbf{IoU75}\\
\hline
Value feature   & 21.9 & 23.2 & 48.9 & 53.1 & 76.0 & 55.2  \\
Key feature (Ours)  & \textbf{24.8} & \textbf{25.7} & \textbf{55.4} & \textbf{58.4} & \textbf{79.2} & \textbf{60.6} \\
\hline
\end{tabular}%
}
\vspace{-0.2cm}
\caption{Ablation study on different feature fusion strategies.}
\vspace{-0.6cm}
\label{tab:gce_fusion}
\end{table}

\section{Conclusions and Limitations} \label{sec:conclusion}
\vspace{-0.1cm}
We propose GCE-Pose, a category-level pose estimation method that leverages category-specific shape and semantic priors to improve pose learning. Our approach achieves state-of-the-art results on two challenging datasets through a Semantic Shape Reconstruction (SSR) module and a Global Context Enhanced (GCE) feature fusion module. GCE-Pose reconstructs partial inputs into complete semantic shapes, providing robust prior guidance for pose estimation. Although our semantic prior is limited to specific categories without scaling to instance-level alignment, our semantic shape reconstruction strategy still performs accurate pose estimation on category-level real data, even under occlusions and sensor artifacts, moving towards application-ready accuracy in category-level pose estimation.

{
    \small
    \bibliographystyle{ieeenat_fullname}
    \bibliography{main}
}

\clearpage
\setcounter{page}{1}
\maketitlesupplementary
\appendix
\renewcommand\thesection{\arabic{section}}
\section{More Implementation Details}
\label{sup:imp_details}
\paragraph{Robust Partial Feature Extraction.} We document more implementation details about GCE-Pose. For feature extraction, we employ DINOv2 \cite{oquab2023dinov2} to process images cropped to 224 $\times$ 224 resolution using the $dinov2\_vits14$ model variant. Point cloud features are extracted via PointNet++~\cite{qi2017pointnetplusplus} with multi-scale grouping, generating per-point features for partial observation and shape reconstruction tasks. Following AG-Pose \cite{lin2024instance}, we utilize 96 key points. For the object-aware chamfer distance, we set the outlier threshold $\tau_1 = 0.1$ in Eq. (1) and the keypoint diversity regularization threshold $\tau_2 = 0.2$ in Eq. (3).

\paragraph{Semantic Shape Reconstruction.} For each model $k$, we initialize prototypes $c^k \in \mathbb{R}^{N \times 3}$ using the K-Means++ algorithm \cite{kmeans}. The deformation field $v^k$ applied to prototype $c^k$ uses point-wise parameterization with vectors of dimension $D \times(N \times 3)$. To balance complexity and efficiency~\cite{loiseau2021representing}, we set the number of basis vectors $D$ to 5.
The deformation network $\mathcal{D}^k$ processes centered partial point clouds to produce shape parameters $a \in \mathbb{R}^{5}$, corresponding to coordinates in the linear space defined in Eq. (4). Similarly, the scale network $\mathcal{S}^k$ outputs scaling parameters $s \in \mathbb{R}^3$ for anisometric scaling along all axes. Both networks share a PointNet++-based \cite{qi2017pointnetplusplus} encoder for feature extraction.
Training proceeds in two stages:
\begin{itemize}
    \item First, we train the deep linear shape model using ground truth point clouds sampled from object meshes via farthest point sampling (FPS). Following \cite{loiseau2021representing}, we employ curriculum learning by optimizing the prototype $c^k$ first, then gradually increasing the deformation field $v^k$ basis vector dimensions. We train for 1000 epochs using the Adam optimizer with a $1e^{-3}$ learning rate. 
    \item Second, we augment centered partial observation inputs with random rotations (0° to 20° on all axes with 0.5 probability). To balance the parameter loss $\mathcal{L}_{\text{para}}$ defined in Eq. (6), we set $\lambda_1 = 1.0$ and $\lambda_2 = 0.1$. For the reconstruction loss in Eq. (7), we use $\lambda_{\text{CD}} = 1.0$ and $\lambda_{\text{para}} = 0.1$. Training runs for 30 epochs using Adam with learning rate $1e^{-3}$.
\end{itemize}

We use Pytorch3D to position 8 virtual perspective cameras in a cube configuration around the normalized target object for semantic prototype construction. We also position a point light to enhance the surface detail and generate realistic rendering. We employ the DINOv2 pre-trained model for feature extraction to generate pixel-aligned feature descriptors. We gathered 200 nearest neighbors for each sampled point using the KNN algorithm described in Eq. (9) to aggregate semantic features from the dense semantic point to our deep linear shape reconstruction.

\paragraph{Pose Size Estimator.}
To train the pose estimation network, we balance the loss function defined in Eq. (18) with hyperparameters: $\lambda_1 = 2.0$, $\lambda_2 = 2.0$, $\lambda_3 = 15.0$, $\lambda_4 = 0.3$, $\lambda_5 = 0.3$. We train the network using ADAM optimizer with CyclicLR scheduler in triangular2 mode with base learning rate $lr = 2e^{-5}$ and max learning rate $lr = 5e^{-4}$. To deal with the symmetry issue, we follow~\cite{tian2020shape} to transform the rotation to canonical. 

\noindent \textbf{Instance-Segmentation.} We follow the previous literature~\cite{lin2023vi, lin2024instance, query6dof, lin2022category, chen2024secondpose} of category-level pose estimation for fair comparisons, using the same segmentation mask as the baseline methods. Specifically, the provided Mask-RCNN segmentation results are used in REAL275, and the GT segmentation mask is used for testing HouseCat6D.

\noindent \textbf{Training process details.} The training process of SSR module is in two stages, where the first stage is used to generate the shape parameters as supervision signal for the second stage, in the second stage, the network is trained with noisy sensor point cloud to make shape reconstruction network robust against noise. The pose estimation part is trained independently after both of these stages.

\section{Evaluation on Instance Reconstruction}
We evaluate our instance reconstruction with chamfer distance (CD) on the HouseCat6D Dataset. We measure the Chamfer distance between our reconstructed pointclouds and the ground-truth pointclouds sampled from the CAD model in NOCS space. We represent the CD metric with
\begin{equation}
\small
d_{\mathrm{CD}}(M, \hat{M})=\sum_{x \in M} \min _{y \in \hat{M}}\|x-y\|^2+\sum_{y \in \hat{M}} \min _{x \in M}\|x-y\|^2 .
\end{equation}
where $M$ and $\hat{M}$ denote the reconstructed point cloud and the ground-truth point cloud sampled from the CAD model, respectively. The term \(\|\cdot\|^2\) represents the squared Euclidean distance, and \(\min\) computes the nearest neighbor distance for each point in one point cloud to the other.

As shown in \cref{tab:rec_housecat}, we achieve $2.39 \times 10^{-3}$ mean chamfer distance of our reconstruction. 
\label{sup:ssr}
\begin{table*}[ht]
\centering

\begin{tabular}{ccccccccccc|c}
\hline
\textbf{Category} & Bottle & Box & Can & Cup & Remote & Teapot & Cutlery & Glass & Tube & Shoe & \textbf{Average} \\
\hline
\textbf{Chamfer Distance} & 1.59   & 7.79 & 3.45 & 1.77 & 1.18  & 2.79 & 0.46 & 1.93  & 1.26 & 1.70 & 2.39 \\
\hline
\end{tabular}
\caption{Reconstruction performance for the categories in HouseCat6D dataset \cite{jung2024housecat6d}. Evaluated with Chamfer Distance metric $\left(10^{-3}\right)$. }
\label{tab:rec_housecat}
\end{table*}

To evaluate the reconstruction performance of our method, we compare the reconstructed shape with the groundtruth shape using the Chamfer Distance. We report the per-category shape reconstruction result in \cref{tab:rec_housecat}.

\section{More Ablation Studies}
\label{sup:more_ablation}
In our Global Context Enhancement Feature Fusion module, we demonstrate the best result using the DINO value feature from partial observation and the DINO key feature from semantic global reconstruction. For symmetric objects, e.g., “glass", as shown in Figure 10, the value features are symmetric and ambiguous around the rotational axis, while the key features are embedded with positional code and thus distinctive, which is helpful when handling symmetric cases. We also report the quantitative results for “glass" in \cref{tab:glass_feature}, showing higher performance when using key features.

Experimental results of the full benchmark on the efficacy of global context feature fusion in \cref{tab:supp_featurefusion} show that our feature fusion strategy can enhance pose estimation performance effectively. 

We additionally conduct experiments on the robustness to hyper-parameter variation. \cref{tab:hyperpara_left} shows the results of hyper-parameter testing, including the number of key points and the hyper-parameters of the loss function defined in Eq. (18). The results are slightly different but still stable. 

\begin{figure}[t]
    \centering
    \includegraphics[width=0.5\textwidth]{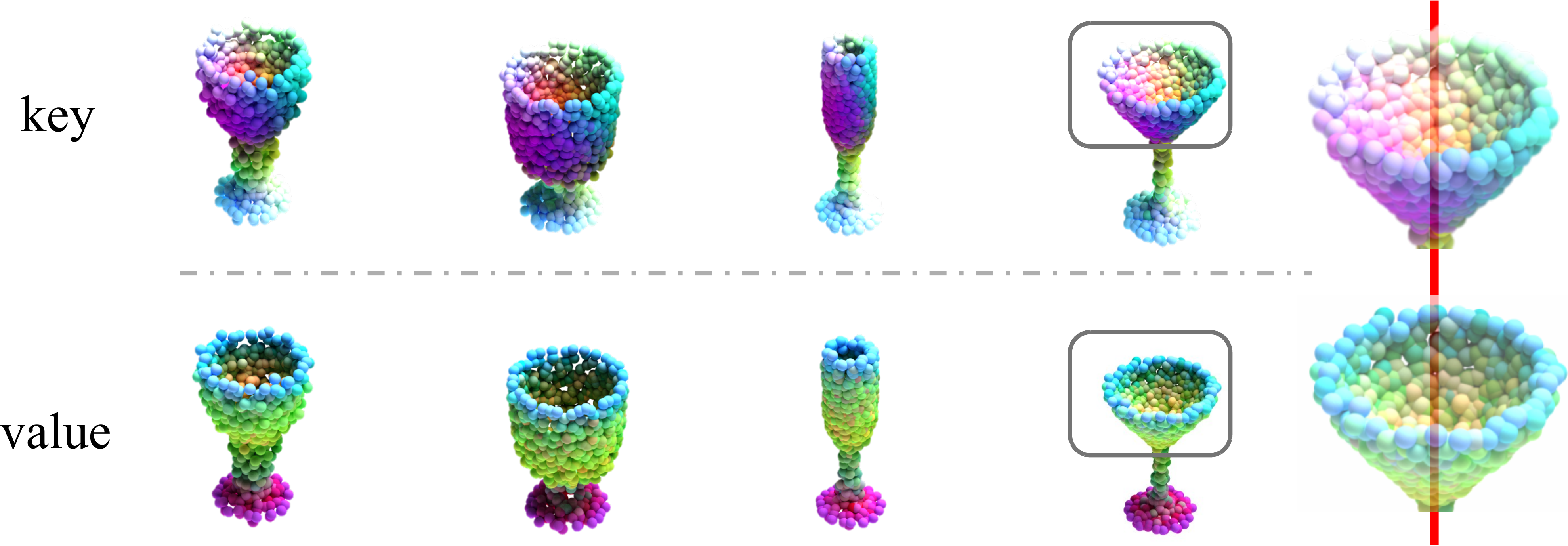}
    \caption{ \footnotesize Visualization of feature point cloud using PCA. Upper row: Key feature. Bottom row: Value feature. The zoom-in visualizations indicate embedding changes in the key feature around symmetric areas, which are negligible for the value feature.}
    \label{fig:vis_feature}   
\end{figure}

\begin{table}[t]
    \centering
    \renewcommand{\arraystretch}{1.2}
    \resizebox{1.0\linewidth}{!}{
    \begin{tabular}{c|c|c|c|c|c}
        \hline
        Keypoint & $\lambda_2$ & $\lambda_3$ & $\mathbf{5^{\circ} 2 cm}$ & $\mathbf{10^{\circ} 2 cm}$ & IoU75 \\
        \hline
        \textbf{96} & \textbf{2.0} & \textbf{15.0} & \textbf{24.8}  & $\mathbf{55.4}$ &  $\mathbf{60.6}$ \\
        128 & 2.0 & 15.0  &  $24.2$  & $52.9$ &  $57.8$ \\
        96  & 0.5 & 15.0 & ${23.5}$  & ${54.6}$ & ${58.9}$ \\
        96 & 2.0 & 3.0 & $23.8$  & ${53.4}$ & ${59.7}$ \\
        \hline
    \end{tabular}
    }
    \caption{Hyperparameter comparison, the default setting is in bold.}
    \label{tab:hyperpara_left}
\end{table}

\begin{table}[t]
    \centering
    \renewcommand{\arraystretch}{1.2}
    \resizebox{1.0\linewidth}{!}{
        \begin{tabular}{c|c|c|c}
            \hline
            {Metrics} & {Value feature} & {Key feature (Ours)} & {Difference} \\
            \hline
            ${5^{\circ} 2 \text{cm}}$ & 59.16 & \textbf{64.86} & 5.70  \\
            ${5^{\circ} 5 \text{cm}}$ & 62.50 & \textbf{66.11}  &  3.61   \\
            ${10^{\circ} 2 \text{cm}}$  & 86.29 & \textbf{92.30} & 6.01  \\
            ${10^{\circ} 5 \text{cm}}$ & 91.72 & \textbf{94.49} & 2.77 \\
            \hline
        \end{tabular}
    }
    \caption{Quantitative comparison for symmetric category ``glass".}
    \label{tab:glass_feature}
\end{table}

\begin{table*}[ht]
    \centering
    \resizebox{\linewidth}{!}{
    \begin{tabular}{l|c c c c | c c c c c c c c}
        \hline
        Method & Ins. recon. & Mean shape & Geo. & Sem. & $5^\circ2$cm & $5^\circ5$cm & $10^\circ2$cm & $10^\circ5$cm & IoU50 & IoU75 \\
        \hline
        (0) AG-Pose (DINO) & \(\times\) & \(\times\)& \(\times\) & \(\times\) &   21.34 & 22.27 & 52.00 & 55.12 & 76.79 & 56.07\\
        (1) Ours (Only Geo.) & \(\checkmark\) & \(\times\)& \(\checkmark\) & \(\times\)& 22.73 & 24.28 & 52.83 & 56.51 & 78.59 & 58.17\\
        (2) Ours (Only Sem.) & \(\times\) & \(\checkmark\)& \(\times\) & \(\checkmark\)& 22.16 & 23.65 & 52.44 & 57.31 & 78.10 & 55.07  \\
        (3) Ours (Mean shape, Geo. \& Sem.) & \(\times\) & \(\checkmark\)& \(\checkmark\) & \(\checkmark\)&  23.37 & 24.24 & 53.85 & 56.83 & 79.27 & 60.46 \\
        (4) GCE-Pose (full pipeline) & \(\checkmark\) & \(\checkmark\)& \(\checkmark\) & \(\checkmark\) &  24.85 & 25.73 & 55.44 & 58.43 & 79.15 & 60.61 \\
        \hline
    \end{tabular}
    }
    \caption{Ablation study on different global priors.
    Ins. recon: instance shape reconstruction as the prior; Mean shape: mean shape of the categories as the prior; Geo.: geometric features from the global prior; Sem.: semantic features for the global prior. 
    }
    \label{tab:supp_featurefusion}
\end{table*}

\begin{table*}[h]
\centering
\label{tab:comp_shape_prior}
\begin{tabular}{l|l|cc|cccc|cc}
\hline
Dataset & Method & Shape Prior & Semantic Prior & $5^\circ2$cm & $5^\circ5$cm & $10^\circ2$cm & $10^\circ5$cm & IoU50 & IoU75 \\
\hline
\multirow{2}{*}{HouseCat6D} & Self-DPDN \cite{lin2022category} & \checkmark & & 6.4 & 6.9 & 22.2 & 25.8 & 56.2 & 26.0 \\
& GCE-Pose (Ours) & \checkmark & \checkmark & 24.8 & 25.7 & 55.4 & 58.4 & 79.2 & 60.6 \\
\hline
\end{tabular}
\caption{Quantitative comparison of category-level object pose estimation with shape-prior on the HouseCat6D dataset \cite{jung2024housecat6d}.}
\end{table*}

\section{More Results and Visualization}
\label{sup:more_housecat}
\begin{figure}[h]
    \centering
    \includegraphics[width=0.5\textwidth]{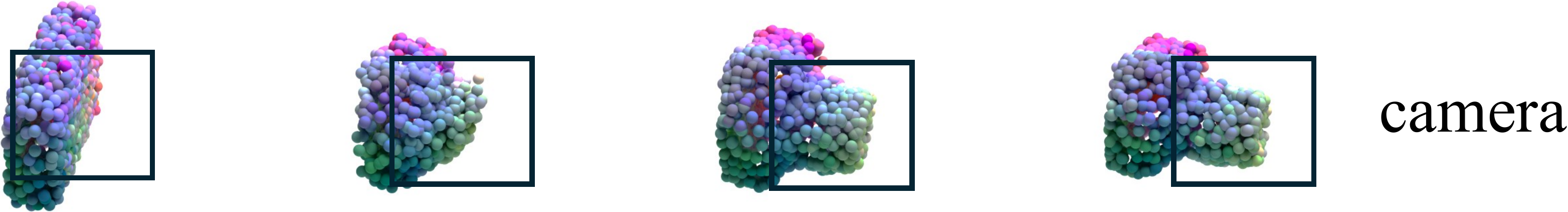}
    \caption{Semantics consistency for the "camera" class despite geometry changes in the lens area.
 }
    \label{fig:camera_semantics}   
\end{figure}

\noindent \textbf{Intra-class semantic variation.} Our method extracts semantic features from the powerful pre-trained large foundational model DINOv2, which is capable of handling intra-class semantic variation effectively as shown in \cref{fig:camera_semantics}. On the other hand, our SSR module is designed to aggregate the categorical semantics into the reconstructed instances effectively through the learned deep linear shapes. As shown in Fig 5 (B) of the main text, with our SSR module, the semantics are consistent across the instances with shape variance demonstrated in \cref{fig:camera_semantics}.

\begin{figure*}[h]
 \centering
    \includegraphics[width=\linewidth]{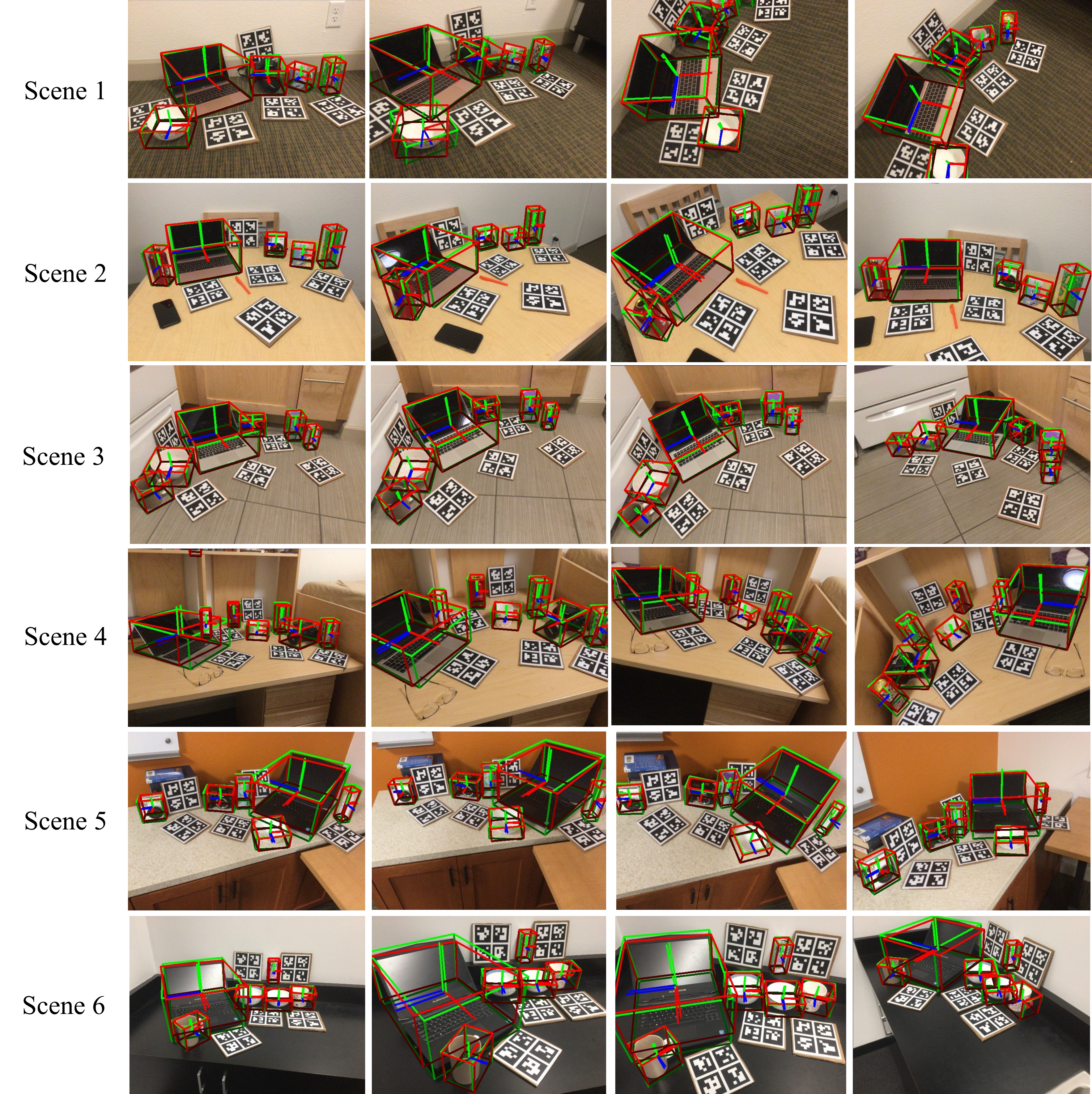}
    \caption{NOCS dataset bounding box visualization. Green  indicates GT, and red indicates prediction results.}
    \label{fig:nocs_bbox_vis}
\end{figure*}

\begin{figure*}[h]
 \centering
    \includegraphics[width=\linewidth]{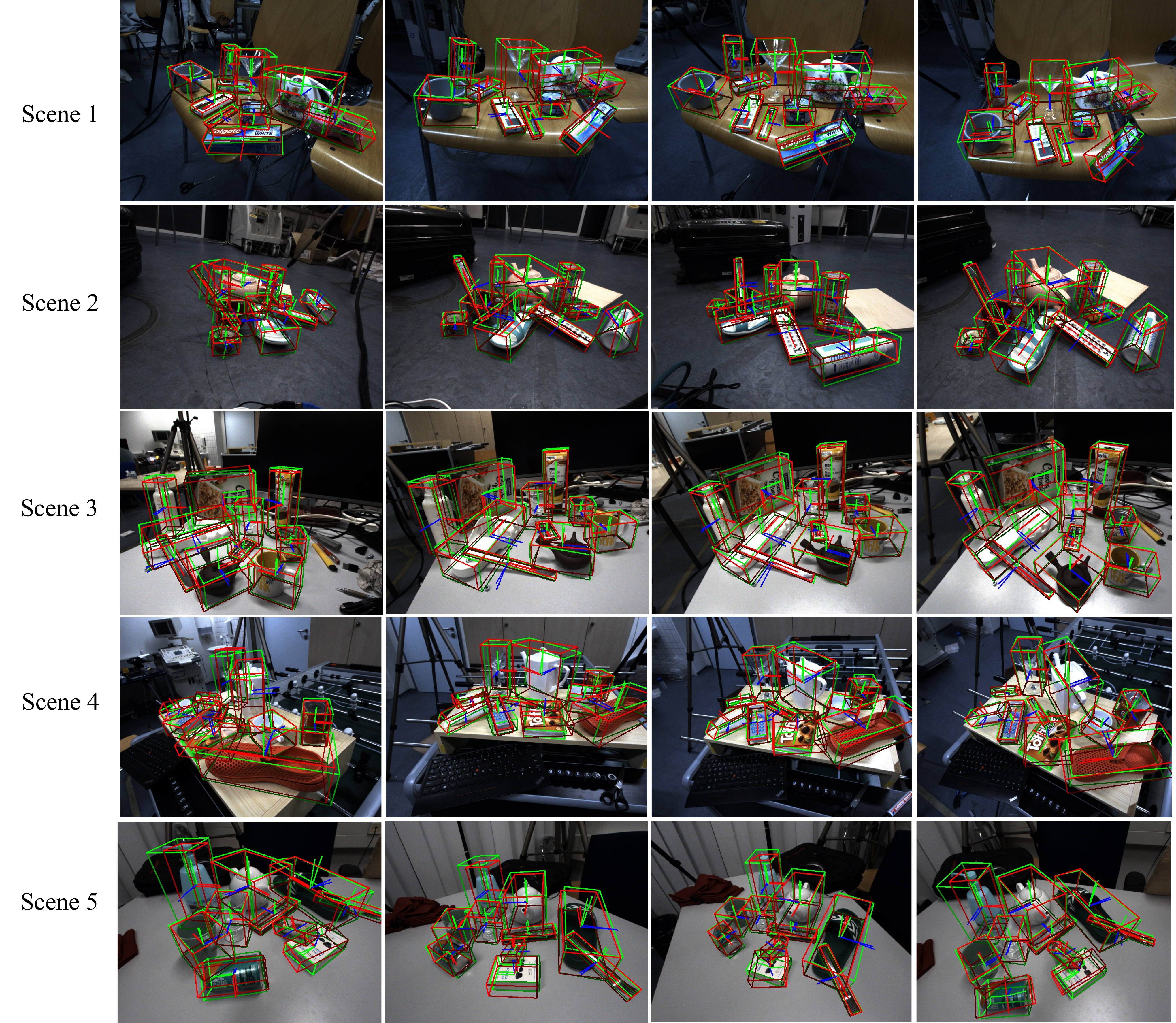}
    \caption{HouseCat6D bounding box visualization. Green  indicates GT, and red indicates prediction results.}
    \label{fig:housecat_bbox_vis}
\end{figure*}

\begin{figure*}[h]
 \centering
    \includegraphics[width=\linewidth]{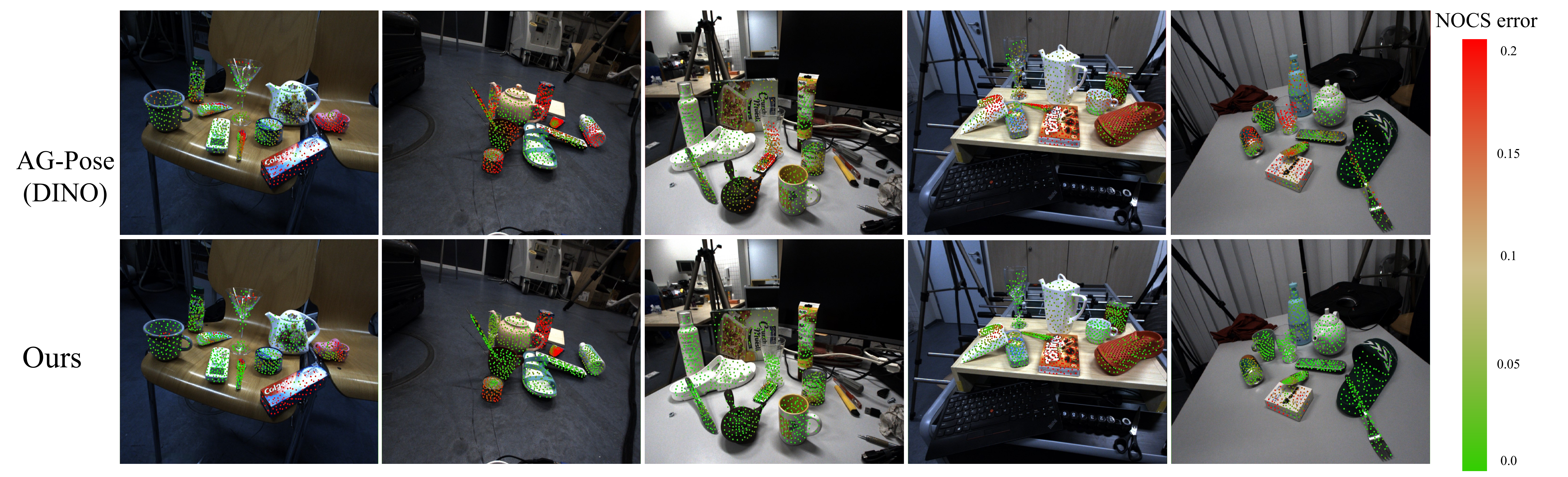}
    \caption{Visualization of HouseCat6D Keypoint NOCS Error. Red indicates a high error; green indicates a low error. }
    \label{fig:housecat_nocs_err}
\end{figure*}

\begin{figure*}[h]
 \centering
    \includegraphics[width=\linewidth]{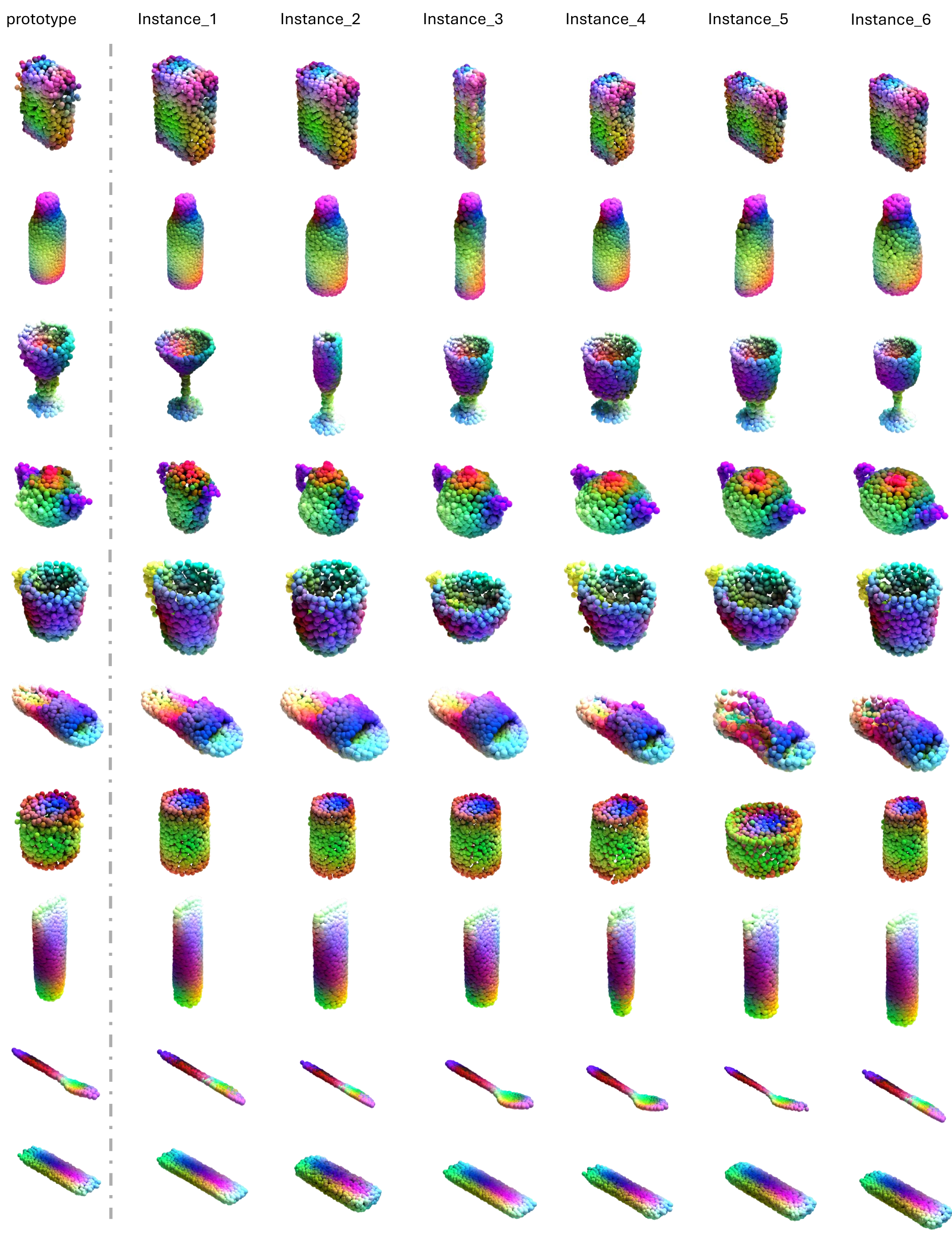}
    \caption{Visualization of Semantic prototypes and in-class semantic transfer results in HouseCat6D dataset.}
    \label{fig:housecat_sem_global}
\end{figure*}

We visualize more results of 3D bounding box prediction of our GCE-Pose for the HouseCat6d dataset in \cref{fig:housecat_bbox_vis} and NOCS dataset in \cref{fig:nocs_bbox_vis}. We choose four images per scene in the test set and indicate the groundtruth results in green and the prediction in red.

To demonstrate the effectiveness of our method in improving pose estimation and NOCS coordinates prediction, we provide further visualization of 3D bounding box prediction and evaluate NOCS errors on key points. \cref{fig:housecat_nocs_err} presents a qualitative comparison of AG-Pose \cite{lin2024instance} (DINO) and our proposed method on the Housecat6D dataset. The visualizations display the NOCS error map overlaid on the images, where red dots indicate higher errors and green dots indicate lower errors. Our qualitative results show that our method achieves high precision in predicting NOCS coordinates, which is essential for accurate pose estimation. Additionally, we have included videos in the attached file that showcase the complete results across the full sequences of the HouseCat6D dataset.

Furthermore, we conducted experiments on HouseCat6D using the state-of-the-art shape-prior-based method, Seld-DPDN~\cite{lin2022category}. \cref{tab:comp_shape_prior} highlights the quantitative results, demonstrating the advantages of incorporating shape and semantic priors for category-level object pose estimation.

We showcase semantic prototype and their corresponding semantic transfers with more classes in the HouseCat6D dataset, as shown in \cref{fig:housecat_sem_global}. The classes, listed from top to bottom, include box, bottle, glass, teapot, cup, shoe, can, tube, cutlery, and remote.

\end{document}